\documentclass[lettersize,journal]{IEEEtran}
\usepackage{amsmath,amsfonts}
\usepackage{algorithmic}
\usepackage{algorithm}
\usepackage{array}
\usepackage[caption=false,font=normalsize,labelfont=sf,textfont=sf]{subfig}
\usepackage{textcomp}
\usepackage{stfloats}
\usepackage{url}
\usepackage{verbatim}
\usepackage{graphicx}
\usepackage{cite}
\usepackage{hyperref}       
\usepackage{url}            
\usepackage{booktabs}       
\usepackage{amsfonts}       
\usepackage{nicefrac}       
\usepackage{microtype}      
\usepackage{xcolor}         
\usepackage{amsmath}
\usepackage{amssymb}
\usepackage{mathtools}
\usepackage{amsthm}
\usepackage{multirow}
\usepackage{wrapfig}
\usepackage{natbib}
\usepackage{microtype}
\usepackage{times}  
\usepackage{helvet}  
\usepackage{courier}  
\usepackage{graphicx} 
\usepackage{natbib}  
\usepackage{caption} 
\usepackage{algorithm}
\usepackage{algorithmic}
\usepackage{amsmath}
\usepackage{multirow}
\usepackage{amssymb}
\usepackage{booktabs}
\usepackage{bbding}
\usepackage{pifont}
\usepackage{wasysym}
\usepackage{amssymb}
\usepackage[table]{xcolor}
\usepackage{newfloat}
\usepackage{listings}
\usepackage{booktabs} 

\usepackage{hyperref}
\usepackage{amsmath}
\usepackage{amssymb}
\usepackage{mathtools}
\usepackage{amsthm}

\usepackage[capitalize,noabbrev]{cleveref}

\theoremstyle{plain}

\theoremstyle{definition}

\theoremstyle{remark}

\begin{document}

\title{Scale Contrastive Learning with Selective Attentions for Blind Image Quality Assessment}

\author{{Runze Hu, Zihao Huang, Xudong Li, Bohan Fu, Yan Zhang, Sicheng Zhao}
\thanks{Runze Hu, Zihao Huang and Bohan Fu are with the School of Information
and Electronics, Beijing Institute of Technology, Beijing 100000, China
(e-mail: hrzlpk2015@gmail.com; 3120240675@bit.edu.cn).}
\thanks{Xudong Li and Yan Zhang are with the School of Key Laboratory of Multimedia Trusted Perception and Efficient Computing, Xiamen University, Xiamen 361005, China (e-mail: timingao@stu.xmu.edu.cn; bzhy986@gmail.com).}
\thanks{Sicheng Zhao is with Beijing National Research Center for Information
Science and Technology (BNRist), Tsinghua University, Beijing 100084,
China (e-mail: schzhao@gmail.com).}
\thanks{Corresponding author: Yan Zhang.}}

\markboth{Journal of \LaTeX\ Class Files,~Vol.~14, No.~8, August~2021}%
{Shell \MakeLowercase{\textit{et al.}}: A Sample Article Using IEEEtran.cls for IEEE Journals}

\IEEEpubid{}

\maketitle

\begin{abstract}

Human visual perception naturally evaluates image quality across multiple scales, a hierarchical process that existing blind image quality assessment (BIQA) algorithms struggle to replicate effectively. This limitation stems from a fundamental misunderstanding: current multi-scale approaches fail to recognize that quality perception varies dramatically between scales—what appears degraded when viewed closely may look acceptable from a distance. This inconsistency not only creates misleading ``visual illusions'' during feature fusion but also introduces substantial redundant information that dilutes quality-critical features and leads to imprecise assessments. Our CSFIQA framework advances multi-scale BIQA via two key innovations: (1) a selective focus attention mechanism that mimics human visual attention by filtering out redundant cross-scale information that would otherwise mask subtle quality indicators, and (2) a scale contrastive learning strategy that explicitly learns to distinguish quality variations both across and within scales. By incorporating an adaptive noise sample matching mechanism, CSFIQA effectively identifies perceptual quality discrepancies in the same content viewed at different scales. Experiments demonstrate substantial improvements over state-of-the-art methods across seven datasets, achieving up to {8.8\%} SRCC improvement on challenging real-world distortions, confirming CSFIQA's superior alignment with human quality perception. 

\end{abstract}


\section{Introduction}
\label{sec:intro}
\begin{figure}[t!]
  \centering
    \includegraphics[width=0.47\textwidth]
    {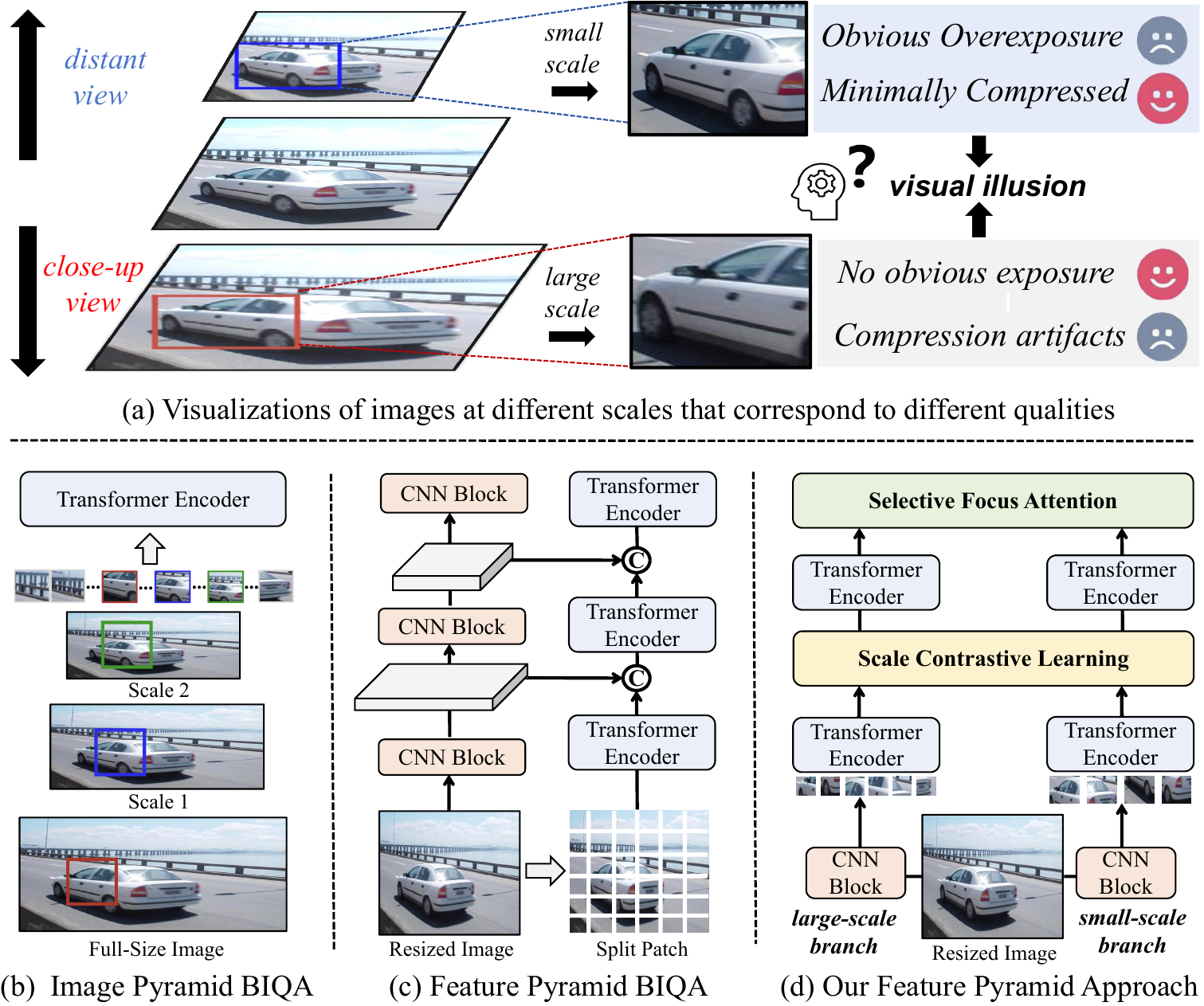}
  \caption{(a) The same image under different patch perspectives can lead to varying quality judgments, and simply combining information from different viewpoints is prone to causing visual hallucinations.
 (b-d) Comparison of mainstream multi-scale paradigms with our approach, which uses scale contrastive learning to distinguish quality differences in (a). The designed selective focus attention module can remove redundant semantic information and enhance attention related to perceptual quality.}
  \label{fig1}
  \vspace{-10pt}
\end{figure}

Image Quality Assessment (IQA) aims to model the human visual system's ability to perceive image quality~\cite{yang2022maniqa,zhang2022perceptual}. It has been widely applied in fields such as image restoration~\cite{zhang2022exploring}, compression, and generation, with the goal of enhancing human visual experience. Based on whether distortion-free reference images are required, IQA can be classified into three types: full-reference, reduced-reference, and no-reference (or blind) IQA. Among these, Blind Image Quality Assessment (BIQA) methods have received increasing attention due to their broad applicability.

However, BIQA faces a fundamental yet largely overlooked challenge: the dramatic variation in quality perception across different scales of the same image. While humans effortlessly integrate these multi-scale impressions, current BIQA algorithms struggle with this inherent complexity, creating a significant gap between algorithmic assessments and human judgment. This perceptual discrepancy severely limits the effectiveness of BIQA in critical applications ranging from image compression and restoration to content generation systems.
Current multi-scale BIQA methods, whether operating at the image level through direct resizing (Fig.~\ref{fig1}(b)) or at the feature level through pyramid structures (Fig.~\ref{fig1}(c)), encounter two critical bottlenecks that fundamentally undermine their performance:

First, the ``visual illusions'' problem presents a severe limitation in existing approaches. Traditional methods erroneously assume that image regions maintain consistent quality characteristics regardless of scale—a fundamentally flawed assumption illustrated in Fig.~\ref{fig1}(a). In reality, quality perception varies dramatically across scales: compression artifacts that dominate perception at a large scale (close-up view) may become imperceptible at smaller scales (distant view), while global distortions like overexposure might dominate at smaller scales but remain unnoticed in large-scale patches. When algorithms naively combine features from multiple scales, they generate misleading quality representations where high-quality and low-quality signals from different scales incorrectly neutralize each other. This creates a perceptual distortion we term the ``visual illusions'' effect—where the algorithm perceives uniform quality across an image despite significant scale-dependent variations—leading to assessments that fundamentally contradict human perception. 

\begin{figure}[t!] 
    \centering 
    \includegraphics[width=0.47\textwidth]{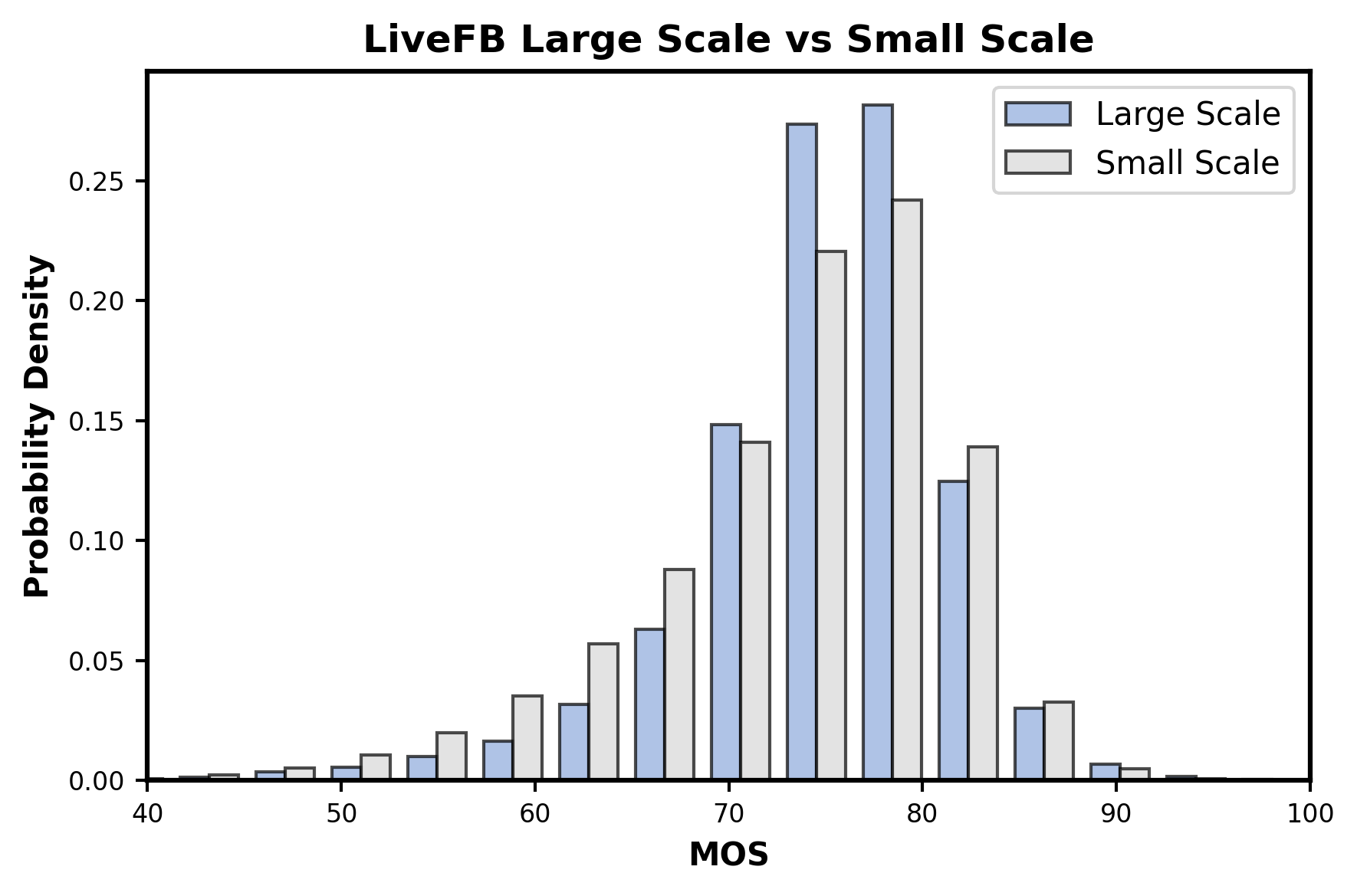} 
    \vspace{-10pt}
    \caption{The MOS distribution of LiveFB in large scale (40\% of the original image size) and small scale (20\% of the original image size).}
\label{mos}
\vspace{-10pt}
\end{figure}

To support this claim, Fig.\ref{mos} presents the Mean Opinion Score (MOS) distributions of the LiveFB dataset at different scales, clearly demonstrating the existence of scale-sensitive perceptual differences. Building upon this observation, Tab.\ref{mos_setting} further shows that neglecting such scale-specific variations during training—by assigning uniform quality labels across scales—can degrade model performance, thereby highlighting the necessity of incorporating scale-aware modeling in IQA.
In addition, our analysis of feature activation maps (Fig.~\ref{visual}) provides a more intuitive visualization of how this illusion effect manifests in practice: current models tend to focus on undistorted regions while neglecting areas that are more indicative of actual image quality degradation.

Second, the ``information dilution'' problem significantly impairs quality detection sensitivity. Unlike human vision, which selectively focuses on quality-relevant features while filtering redundant information, current multi-scale methods indiscriminately process all cross-scale information. This approach not only wastes computational resources but, more critically, creates a signal-to-noise problem where quality-critical features become overwhelmed by redundant semantic content shared across scales. As demonstrated in our visualization results (Fig.~\ref{visual}), this dilution effect causes quality indicators—particularly subtle artifacts and distortions—to be masked by dominant semantic features that remain consistent across scales but contribute little to quality assessment. This effect is especially pronounced in authentic distortion datasets where quality variations are complex and multifaceted.

To address these fundamental limitations, we propose CSFIQA (Contrast-Constrained Scale-Focused Image Quality Assessment), a novel framework specifically designed to accurately model scale-dependent quality perception. Our approach introduces two complementary innovations:

The first innovation directly tackles the ``information dilution'' problem through a Selective Focus Attention (SFA) mechanism. This mechanism intelligently identifies and filters redundant cross-scale information by preserving only the most relevant attention values through an adaptive filtering selector. It then amplifies quality-discriminative features through an information concentrator module, effectively mimicking the human visual system's ability to focus attention on perceptually important regions while suppressing irrelevant information. This approach significantly enhances the model's ability to isolate and emphasize quality-critical features that would otherwise be overwhelmed by redundant semantic content.

The second innovation addresses the ``visual illusions'' problem through a comprehensive Scale Contrastive Learning (SCL) framework with an adaptive Noise Sample Matching (NSM) mechanism. This approach explicitly teaches the model to distinguish between different quality characteristics at both inter-scale (across different scales) and intra-scale (within the same scale) levels. Crucially, the NSM mechanism specifically targets regions with inconsistent quality information across scales within the same image, enabling the model to accurately represent scale-dependent quality variations rather than incorrectly averaging them. This explicit modeling of scale-quality relationships effectively prevents the visual illusion effect that plagues traditional multi-scale approaches.

Our comprehensive evaluation on seven benchmark datasets demonstrates that CSFIQA consistently outperforms state-of-the-art methods, with particularly significant improvements on challenging real-world datasets: 8.8\% SRCC improvement on LIVEFB and 1.6\% on LIVEC. These results confirm that by accurately addressing the fundamental limitations of scale-dependent quality perception, our approach successfully bridges the gap between algorithmic assessment and human perception of image quality.

\section{Related Work}
\label{section2}

\begin{figure*}[t!]
  \centering
\includegraphics[width=0.95\textwidth]{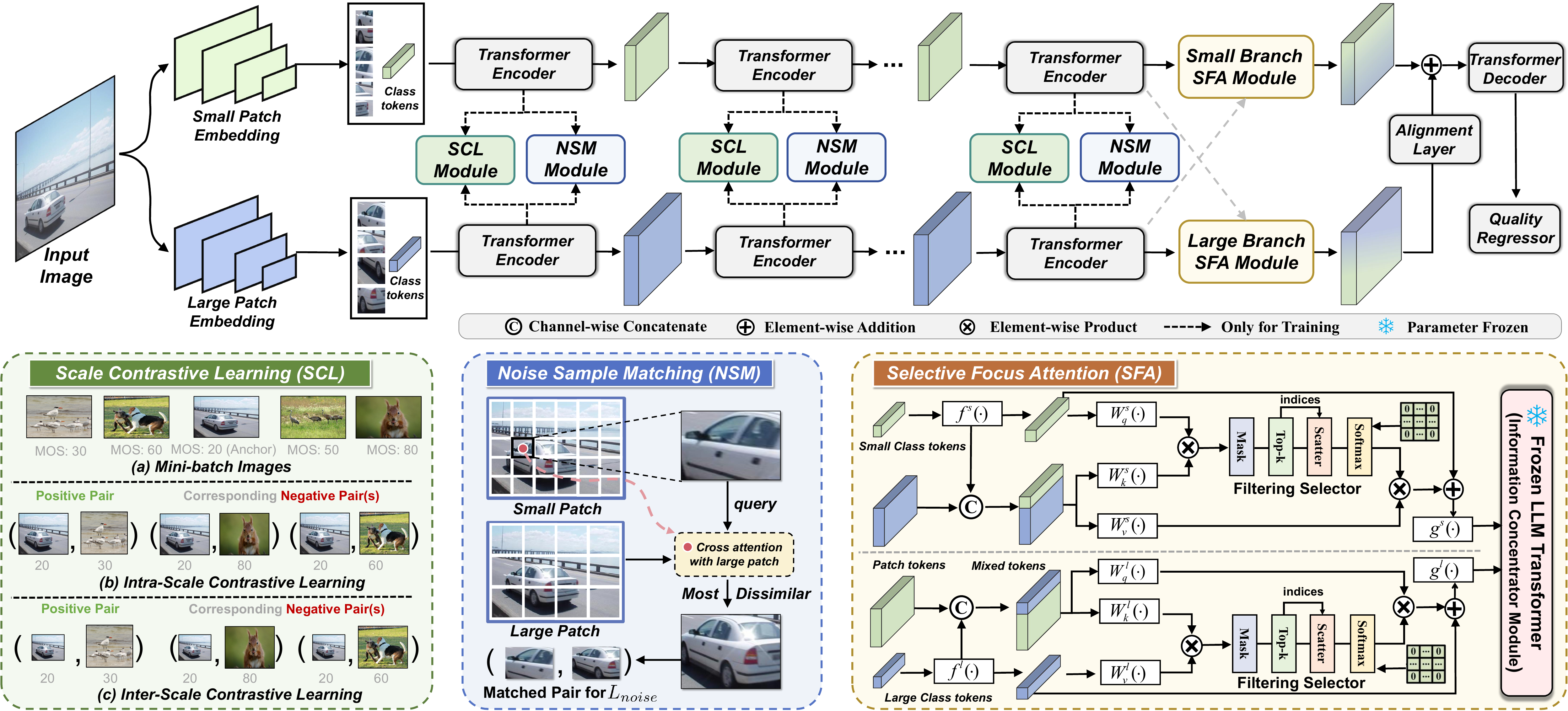}
  \caption{ The overall of our CSFIQA. For a given image, the \( l \)-th layer Transformer Encoder extracts image features $F^l_a$ ($a \in \{small, large\}$) at different scales. These features are then input into the SCL module (Sec.~\ref{SCL}), which uses the relative distance of quality labels in the mini-batch to identify inter and intra-scale positive and negative samples for contrastive learning, enhancing the ability to discriminate quality differences between images. The NSM (Sec.~\ref{SCL}) further increases the distance between quality representations that exhibit significant differences across multiple scales within the same image, thereby distinguishing subtle quality variations between regions to address the ``visual illusions'' effect caused by varying quality information (see Fig.~\ref{fig1}(a)). Features from the final encoder layer \( F^L_a \) enter the SFA module (Sec.~\ref{SFA}), where the Adaptive Filtering Selector (AFS) retains the top-k self-attention similarity scores, and the Information Concentrator Module amplifies important details to obtain quality-aware features. These features are then decoded to predict the quality \(\hat{Y}\) (see Algorithm \ref{code:ref} for details).
  }
  \label{framework}
\end{figure*}

\subsection{Multi-Scale and Contrastive Learning in BIQA}
Deep learning has significantly advanced Blind Image Quality Assessment (BIQA), with models evolving from Convolutional Neural Networks (CNNs) to Vision Transformers (ViTs). While CNNs excel at extracting local features~\cite{bosse2017deep}, they struggle to capture long-range dependencies. ViTs and hybrid architectures~\cite{DEIQT, golestaneh2022no} overcome this using self-attention but can introduce parameter redundancy or over-emphasize global semantics at the expense of sensitivity to local distortions. To address these limitations, multi-scale architectures have become central to BIQA, mimicking the human ability to integrate fine details with global context~\cite{su2020blindly, chen2024topiq}. Despite their progress, these methods often fail to effectively model complex, scale-sensitive quality discrepancies and can introduce redundant information, highlighting the need for a more refined mechanism to balance local fidelity and global context.

Alongside architectural evolution, contrastive learning has emerged as a powerful paradigm for learning robust quality representations~\cite{zhao2023quality, liintegrating}. Seminal works like CONTRIQUE~\cite{madhusudana2022image} and Re-IQA~\cite{reiqa} demonstrated its effectiveness. However, these methods typically rely on distortion types or levels as supervisory signals, rather than ground-truth quality scores. They also tend to treat any two images with different content as a negative pair, overlooking the nuanced quality relationships that might exist between them. Our work addresses these gaps by introducing scale contrastive learning, which directly models the relationship between quality at different resolutions and the overall perceptual score.el the relationship between different quality regions and the overall image quality.

\section{Proposed Method}\label{section3}
\subsection{Overall Pipeline}
We proposed CSFIQA, which is designed to precisely model the relationship between image scale and overall quality. As illustrated in Fig.~\ref{framework}, CSFIQA integrates three primary modules: the Scale Contrastive Learning module \textbf{(SCL)}, the Noise Sample Matching module \textbf{(NSM)}, and the Selective Focus Attention module \textbf{(SFA)}. Initially, the input image $I$ is segmented into patches of varying scales, which are  independently processed through distinct $L$ layers of the Transformer encoder to extract scale-level features  $F^l_a$ ($a \in \{small, large\}$) at the $l$-th layer. The scale-level features are subsequently forwarded to the SCL (Sec.~\ref{SCL}) to derive positive and negative samples, denoted as $P$ and $N$. They are utilized to compute their respective InfoNCE loss. Simultaneously, through NSM module (Sec.~\ref{SCL}), we calculate the similarity between image patches with the largest quality gap within the same image as a loss to distinguish them. The output of the final encoder layer is passed to the SFA (Sec.~\ref{SFA}) to filter redundant features and amplify quality features. Finally, we obtain multi-scale quality features, refined via an alignment layer and Transformer decoder to yield predicted quality scores (see Algorithm \ref{code:ref}).
\vspace{-0.2cm}
\subsection{Scale Contrastive Learning}~\label{SCL}
To enhance the model's perception of scale-dependent quality, we introduce Intra- and Inter-Scale contrastive learning. This approach ensures that representations for images of similar quality are consistent across different scales, while representations for images of dissimilar quality are pushed apart. Consequently, the model can better integrate fine-scale features, which are sensitive to local distortions like noise and artifacts, with coarse-scale features that capture global distortions like overexposure and blur. 

As shown in the SCL module (Fig.~\ref{framework}c), we use Mean Opinion Score (MOS) similarity to select positive and negative sample pairs, aligning feature representations with perceptual quality. Specifically, for a given query patch~$i$ and another patch~$j$ in a mini-batch of size~$B$, we start with their MOS vector $y \in \mathbb{R}^B$. 
We then compute a pairwise score distance matrix $Y_d \in \mathbb{R}^{B \times B}$ using the Manhattan distance. 
Let $Y_d^{ij}$ be the distance between the scores of~$i$ and~$j$, and let $Y_{d, \text{max}}^i = \max_{k} Y_d^{ik}$ be the maximum distance in the query's row. 
Using two threshold coefficients, $\gamma_1$ and $\gamma_2$, we define the sample classifier as:
\begin{algorithm}[tb]
\caption{Pseudocode for proposed CSFIQA}
\small 
\begin{algorithmic}[1]
\setlength{\itemsep}{2pt} 
\setlength{\parskip}{1pt}

\STATE \textbf{Input:} Mini-batch Images $I = \{X_i, Y_i\}_{i=1}^{B}$

\STATE \textbf{Variables:}
\begin{ALC@g}
    \STATE $B$: Batch size
    \STATE $F_a^i$: Image feature
    \STATE $P$: Positive pairs
    \STATE $N$: Negative pairs
    \STATE $M$: Small patch regions by NSM
    \STATE $K'$: Neighbouring large patch regions
\end{ALC@g}

\STATE \textbf{Output:} Predicted quality score $\{\hat{Y}\}_{i=1}^{B}$

\FOR{\texttt{l, blk in enumerate(blocks)}}
    \STATE \hspace{-0.5em}$\triangleright$ \textit{Transformer encoder}
    
    \FOR{$i = 1$ to $B$}
        \STATE \hspace{-0.5em}$\triangleright$ \textit{Calculate the $P$ and $N$ pairs by Eq.~3}
        
        \STATE $\mathcal{L}(F_a^{l}, F^{+}) = \log \dfrac{-\exp\left(\frac{F_a^{l} \cdot F^{+}}{\tau}\right)}{\exp\left(\frac{F_a^{l} \cdot F^{+}}{\tau}\right) + \sum_{F^{-} \in N} \exp\left(\frac{F_a^{l} \cdot F^{-}}{\tau}\right)}$
        
        \STATE $\mathcal{L}_{\text{scale}} \mathrel{+}= \mathcal{L}(F_a^{l}, F^{+})$
        
        \FOR{$m = 1$ to $M$}
            \STATE $K' = \text{neighbouring}(m)$
            
            \FOR{$k = 1$ to $K'$}
                \STATE $\text{Sim}(G_{\text{small}}^m, G_{\text{large}}^k) = \dfrac{G_{\text{small}}^m \cdot G_{\text{large}}^k}{\|G_{\text{small}}^m\| \|G_{\text{large}}^k\|}$
                
                \STATE $\mathcal{L}_{\text{noise}} \mathrel{+}= \dfrac{1}{\exp(\text{Sim}(G_{\text{small}}^m, G_{\text{large}}^k))}$
            \ENDFOR
        \ENDFOR
    \ENDFOR
    
    \STATE \hspace{-0.5em}$\triangleright$ \textit{Selective Focus Attention}
    \STATE $F_a = \text{SFA}(F_a^B)$
\ENDFOR

\STATE \hspace{-0.5em}$\triangleright$ \textit{Obtain $F$ from $F_a$ through the Alignment Layer}
\STATE $\hat{Y} = \text{Decoder}(F)$ \hfill $\triangleright$ \textit{Transformer decoder}

\STATE $\mathcal{L}' = \left\| \hat{Y} - Y \right\|_1 + \lambda \left(\mathcal{L}_{\text{scale}} + \mathcal{L}_{\text{noise}}\right)$

\end{algorithmic}
\label{code:ref}
\end{algorithm}

\begin{equation}
Classifier(i, j) \in 
\begin{cases} 
P, & if \;  Y_d^{ij} \leq \gamma_1 * Y_d^{\hat{i}}, \\
N, & if \;  Y_d^{ij}  > (1 - \gamma_2) * Y_d^{\hat{i}}.
\end{cases}
\label{choose}
\end{equation}
Therefore, given a pacth with feature $F_a^{l}\in P$, we define scale-level contrastive loss as follows:
\begin{equation}  
\mathcal{L}_{scale} = \sum_{l=1}^{L} \sum_{a \in \{s,l\}} \frac{1}{|P|} \sum_{F^+\in P} \mathcal{L}(F^{{l}}_a, F^{+}),
\end{equation}
where
\begin{equation}\mathcal{L}(F^{{l}}_a, F^{+}) = \log \frac{-\exp(\frac{F^{{l}}_a \cdot F^{+}} {\tau})}{\exp(\frac{F^{{l}}_a \cdot F^{+}} {\tau}) + \sum_{F^{-} \in N} \exp(\frac{F^{l}_a \cdot F^{-}} {\tau})}.\end{equation}
Here, \( \tau \) denotes the temperature hyperparameter and \( F^{l}_a \) represents the features of patch at any scale in layer $l$ of the encoder. Empirically, we set $\gamma_1$ to 0.2, $\gamma_2$ to 0.7, and $\tau$ to 0.3. Ablation studies on hyperparameter settings are presented in Tab.~\ref{range} and Tab.~\ref{lda}. 

\begin{table*}[ht]
\caption{Performance comparison based on average SRCC and PLCC. Bold values denote the best and second-best results.} 
\setlength\tabcolsep{0.7pt}
\renewcommand\arraystretch{0.8}
    \centering
    \resizebox{1\textwidth}{!}{
        \begin{tabular}{lcccccc||cccccccc}
      \toprule[1.5pt]
    \multirow{2}{*}{Method} & \multicolumn{2}{c}{LIVE} & \multicolumn{2}{c}{CSIQ} & \multicolumn{2}{c||}{TID2013}  & \multicolumn{2}{c}{LIVEC} & \multicolumn{2}{c}{KonIQ} & \multicolumn{2}{c}{LIVEFB} & \multicolumn{2}{c}{SPAQ}\\
    \cmidrule{2-15}     & \multicolumn{1}{c}{PLCC} & \multicolumn{1}{c}{SRCC}  & \multicolumn{1}{c}{PLCC} & \multicolumn{1}{c}{SRCC}& \multicolumn{1}{c}{PLCC} & \multicolumn{1}{c||}{SRCC}& \multicolumn{1}{c}{PLCC} & \multicolumn{1}{c}{SRCC}& \multicolumn{1}{c}{PLCC} & \multicolumn{1}{c}{SRCC}& \multicolumn{1}{c}{PLCC} & \multicolumn{1}{c}{SRCC}& \multicolumn{1}{c}{PLCC} & \multicolumn{1}{c}{SRCC}\\
    \midrule
    BRISQUE ~\cite{BRISQUE} & 0.944 & 0.929 & 0.748 & 0.812 & 0.571 & 0.626  & 0.629 & 0.629 & 0.685 & 0.681 & 0.341 & 0.303 & 0.817 & 0.809 \\
    ILNIQE ~\cite{ILNIQE} & 0.906 & 0.902 & 0.865 & 0.822 & 0.648 & 0.521  & 0.508 & 0.508 & 0.537 & 0.523 & 0.332 & 0.294 & 0.712 & 0.713 \\
    BIECON ~\cite{BIECON} & 0.961 & 0.958 & 0.823 & 0.815 & 0.762 & 0.717 & 0.613 & 0.613 & 0.654 & 0.651 & 0.428 & 0.407 & {-} & {-} \\
    MEON ~\cite{MEON} & 0.955 & 0.951 & 0.864 & 0.852 & 0.824 & 0.808  & 0.710  & 0.697 & 0.628 & 0.611 & 0.394 & 0.365 & {-} & {-} \\
    DBCNN ~\cite{zhang2018blind} & 0.971 & 0.968 & {0.959} & {0.946} & 0.865 & 0.816  & 0.869 & 0.851 & 0.884 & 0.875 & 0.551 & 0.545 & 0.915 & 0.911 \\
    MetaIQA ~\cite{zhu2020metaiqa} & 0.959 & 0.960  & 0.908 & 0.899 & 0.868 & 0.856  & 0.802 & 0.835 & 0.856 & 0.887 & 0.507 & 0.54  & {-} & {-} \\
    P2P-BM ~\cite{ying2020patches} & 0.958 & 0.959 & 0.902 & 0.899 & 0.856 & 0.862   & 0.842 & 0.844 & 0.885 & 0.872 & 0.598 & 0.526 & {-} & {-} \\
    HyperIQA ~\cite{hypernet} & 0.966 & 0.962 & 0.942 & 0.923 & 0.858 & 0.840  & 0.882 & 0.859 & 0.917 & 0.906 & 0.602 & 0.544 & 0.915 & 0.911 \\
    MUSIQ ~\cite{musiq} & 0.911 & 0.940  & 0.893 & 0.871 & 0.815 & 0.773  & 0.828 & 0.785 & {0.928} & {0.916} & {0.661} & {0.566} & {0.921} & {0.918} \\
    TReS  ~\cite{TReS} & 0.968 & 0.969 & 0.942 & 0.922 & 0.883 & 0.863  & 0.882 & 0.859 & {0.928} & 0.915 & 0.625 & 0.554 & {-} & {-} \\
    DACNN ~\cite{pan2022dacnn} & {0.980}  & {0.978} & {0.957} & {0.943} & {0.889} & {0.871}  & {0.884} & {0.866} & 0.912 & 0.901 & {-} & {-} & {0.921} & 0.915 \\
    Re-IQA ~\cite{reiqa} & {0.971}  & {0.970} & {0.96} &0.947 & {0.861} & {0.804}  &{0.854} &{0.84} &{0.923} & {0.914}  & {-} & {-} & {0.925} & {0.918}\\
    DEIQT ~\citep{DEIQT} & \textbf{0.982} & 0.980 & 0.963 & {0.946} & \textbf{0.908} &{0.892}  & {0.894} & {0.875}& {0.934} & 0.921 & {0.663} & {0.571} & {0.923} & {0.919} \\
    CLIP-IQA+ ~\citep{clipiqa} & - & - & - & - &-  &- &{0.832} & {0.805} & 0.909 & 0.895 & 0.593 & 0.575 & {0.866} & {0.864} \\
    
    CDINet ~\cite{10440553} & {0.975} & {0.977} & {0.960} & {0.952} & \textbf{0.908} & \textbf{0.898}  & {0.880} & 0.865& 0.928 & 0.916 & \textbf{-} & \textbf{-} & 0.922 & 0.919 \\
    QFM-IQM ~\cite{10.5555/3692070.3693183} & \textbf{0.983} & \textbf{0.981} & \textbf{0.965} & \textbf{0.954} & {-} &{-}  & \textbf{0.913} & \textbf{0.891}& \textbf{0.936} & {0.922} & 0.667 & 0.567 & 0.924 & \textbf{0.920} \\
    LoDa ~\cite{xu2024boosting} & {0.979} & {0.975} & {-} & {-} & {0.901} &{0.869}  & {0.899} & 0.876& \textbf{0.944} & \textbf{0.932} & \textbf{0.679} & \textbf{0.578} & \textbf{0.928} & \textbf{0.925} \\
    \midrule
    \rowcolor{gray!20}
    CSFIQA (ours) & \textbf{0.983} & \textbf{0.982} & \textbf{0.973} & \textbf{0.967} & \textbf{0.917} & \textbf{0.899}  & \textbf{0.922} & \textbf{0.905} & \textbf{0.944} & \textbf{0.924} & \textbf{0.701} & \textbf{0.629} & \textbf{0.935} & \textbf{0.925} \\
    \bottomrule
    \end{tabular}}
    
  \label{performance}
  \vspace{-0.2cm}
\end{table*}

\begin{table}[t]
\caption{SRCC on the cross datasets validation. The best performances are highlighted in boldface.}
\setlength\tabcolsep{3.5pt}  
\renewcommand\arraystretch{1.15}  
\small  
  \centering
    \begin{tabular}{ccccccc}
    \toprule
    \textbf{Training} & \multicolumn{2}{c}{\textbf{LIVEFB}} & \multicolumn{1}{c}{\textbf{LIVEC}} & \multicolumn{1}{c}{\textbf{KonIQ}} & \multicolumn{1}{c}{\textbf{LIVE}} & \multicolumn{1}{c}{\textbf{CSIQ}} \\
    \midrule
    \textbf{Testing} & KonIQ & LIVEC & KonIQ & LIVEC & CSIQ & LIVE \\
    \midrule
    DBCNN & 0.716 & 0.724 & 0.754 & 0.755 & 0.758 & 0.877 \\
    P2P-BM & 0.755 & 0.738 & 0.740 & 0.770 & 0.712 & - \\
    TReS  & 0.713 & 0.74  & 0.733 & 0.786 & 0.761 & - \\
    DEIQT & 0.733 & 0.781 & 0.744 & 0.794 & 0.781 & 0.932 \\
    LoDa & 0.763 & \textbf{0.805} & 0.745 & 0.811 & - & - \\
    \midrule
    \rowcolor{gray!25}    
    \textbf{Ours}  & \textbf{0.785} & \textbf{0.805} & \textbf{0.762} & \textbf{0.838} & \textbf{0.786} & \textbf{0.933} \\
    \bottomrule
    \end{tabular}
  \label{crossdata}
\end{table}

\begin{table}[t]
\caption{Ablation experiments on different modules.}
\setlength\tabcolsep{5pt}  
\renewcommand\arraystretch{1.15}  
\small  
\centering
\begin{tabular}{ccc|cc|cc} 
    \toprule
     \multicolumn{3}{c|}{\textbf{Method}} & \multicolumn{2}{c|}{\textbf{LIVEC}} & \multicolumn{2}{c}{\textbf{CSIQ}}  \\ 
    \midrule
    Baseline & SCL & SFA  & PLCC  & SRCC  & PLCC  & SRCC    \\ 
    \midrule
    \CheckmarkBold & & & 0.896 & 0.878 & 0.964 & 0.948 \\
    \rowcolor{gray!25} 
    \CheckmarkBold & \CheckmarkBold & & \textbf{0.911} & \textbf{0.892} & \textbf{0.970} & \textbf{0.963} \\
    \CheckmarkBold & & \CheckmarkBold & 0.904 & 0.887 & 0.965 & 0.956 \\
    \midrule
    \CheckmarkBold & \CheckmarkBold & \CheckmarkBold & \textbf{0.922} & \textbf{0.905} & \textbf{0.973} & \textbf{0.967} \\
    \bottomrule
\end{tabular}
\label{ab_model}
\end{table}

\noindent \textbf{Noise Sample Matching (NSM).}
We propose a simple but effective adaptive noise sample matching mechanism to further distinguish samples with inconsistent quality information across different scales of the same image.  We identify the sample in the neighbouring region at scale $s$ with the least similar quality information as a negative sample for contrastive learning for a given image at scale $l$.

First, we divide the feature maps into regions for different scales. 
We obtain features with \( G_a \in \mathbb{R}^{N_a \times D_a} \) from the patch embedding in the ViT encoder and reorganize these patches into a feature map \( \hat{G}_a \in \mathbb{R}^{H_a \times W_a \times D_a} \) with equal height and width. We then apply a sliding window function \( W_a \in \mathbb{R}^{H'_a \times W'_a} \) to further partition these patches into regions. Simultaneously, based on spatial coordinates, we record the large patch regions that encompass each small pacth region, designating them as neighboring regions corresponding to each small patch region. We define $M$ blocks obtained by partitioning at the $small$ patch and $K$ blocks obtained by partitioning at the $large$ patch, wherein the number of neighboring large patch regions for each small patch is $K^{'} (K^{'} \leq K)$. For each region \( G^m_{small} \) at  \( small \) patch, we compute its cosine similarity with each neighbouring region \( G^{k}_{large} \) at \( large \) patch:
\begin{equation}
\label{sim}
Sim(G^m_{small}, G^{k}_{large}) = \frac{G^m_{small} \cdot G^{k}_{large}}{\|G^m_{small}\| \|G^{k}_{large}\|}.
\end{equation}
We compute the loss across all neighboring regions in the small patch feature map, as shown in Eq.~(\ref{all_sim}). This approach enables us to amplify the distance between samples with inconsistent quality information at different scales within the same image, thereby more effectively differentiating between them.
\begin{equation}
\label{all_sim}
\mathcal{L}_{noise} = \sum_{m=1}^{M}\sum_{k=1}^{K^{'}} \frac{1}{\exp(Sim(G^m_{small}, G^k_{large})}.
\end{equation}
\noindent \textbf{Overall Loss.}
Let \( \hat{Y} \) and \( Y \) respectively denote the predicted scores and the ground truth scores for the image \( I \). Given \( \lambda \) represent the hyperparameters. The notation \( \| \cdot \|_1 \) signifies the \( \ell_1 \) regression loss. The total loss is defined as:
\begin{equation}
\mathcal{L} = \sum_I \left( \left\| \hat{Y} - Y\right\|_1 + \lambda \left(\mathcal{L}_{scale} + \mathcal{L}_{noise}\right)\right).
\label{loss}
\end{equation}

\subsection{Selective Focus Attention}\label{SFA}
\noindent \textbf{Preliminaries.} Traditional multi-scale Image Quality Assessment (IQA) methods often suffer from information redundancy, causing them to overlook critical quality-related features. To address this, we propose the \textbf{Selective Focus Attention (SFA)} module. The SFA consists of two sequential components: an \textbf{Adaptive Filtering Selector (AFS)} and an \textbf{Information Concentrator Module (ICM)}. The AFS first employs a filtering attention mechanism to select the most salient cross-scale information. Subsequently, the ICM, which combines a learnable linear layer with a frozen large language model, refines these selected features to pinpoint quality-specific content, thus reducing redundancy in the process.
We begin by examining the cross-attention mechanism commonly used in multi-scale models. For different branches labeled as large and small, the class token from the large branch is concatenated with the patch token from the small branch:
\begin{eqnarray}
\label{attn}
\begin{split}
x' &= \left[x^{large}_{\text{cls}},x^{small}_{\text{patch}} \right],\\
CrossAtt(x') &= \text{softmax}\left(\frac{Q K^T}{\sqrt{C/h}}\right)V.
\end{split}
\end{eqnarray} 
Here, \( C \) represents the number of channels, and \( h \) denotes the number of heads. Given \( Q = x'_{\text{cls}} W_q \), \( K = x' W_k \), and \( V = x' W_v \), this enables the fusion of cross-scale features. 

\noindent \textbf{Adaptive Filtering Selector (AFS).} 
The core of our approach is the {AFS} mechanism, which enhances the standard attention computation from Eq.~\ref{attn}. Instead of using the full attention matrix, AFS implements an \textbf{adaptive top-k filtering} strategy by applying a learnable masking operator, $\mathcal{M}$, to the raw attention scores. For each query, this operator dynamically selects the top-$k$ most relevant key-value pairs. The value of $k$ is not fixed but is determined by a learnable parameter constrained to a fractional range $[\alpha, \beta]$ of the total tokens. This allows the model to adaptively decide how much information to prune. Attention scores not within the top-$k$ are masked with $-\infty$ before the softmax function, effectively nullifying their contribution. The filtering mechanism is formally defined as:
\begin{equation}
\text{SelectAtt}(Q, K, V) = \text{softmax} \left( \mathcal{M} \left( \frac{{Q} {K}^\top}{\sqrt{d}} \right) \right) V.
\end{equation}
Here, $\mathcal{M}$ is the top-k operator, with $[\alpha, \beta] = [1/3, 3/4]$.

\noindent \textbf{Information Concentrator Module (ICM).} Recent work~\citep{pang2023frozen} has shown that frozen LLM encoders can discern information-rich visual tokens and further enhance their contributions to latent representations. In our approach, we have implemented the filtering of features at the scale level before inputting them into the frozen LLM layer. As a result, the frozen LLM layer functions as a scale information amplifier, exhibiting a stronger focus on the feature content that we consider essential. The specific structure of the Information Concentrator and the visualization results of the SFA will be presented in the Sec.~\ref{sec:SFA_detail}.

\section{Experiments}
\subsection{Datasets and Evaluation Protocols}
We evaluated our model on eight public Image Quality Assessment (IQA) datasets. Four datasets feature authentic distortions: LIVEC~\cite{ghadiyaram2015massive}, KonIQ-10k~\cite{hosu2020koniq}, LIVEFB~\cite{ying2020patches}, and SPAQ~\cite{fang2020perceptual}. The other four contain synthetic distortions: LIVE~\cite{sheikh2006statistical}, CSIQ~\cite{larson2010most}, TID2013~\cite{ponomarenko2015image}, and KADID~\cite{lin2019kadid}. These datasets vary significantly in scale and content, from hundreds of images with a few distortion types to nearly 40,000 images with diverse artifacts.
To measure performance, we used the Spearman Rank Correlation Coefficient (SRCC) and the Pearson Linear Correlation Coefficient (PLCC). For both metrics, values closer to 1 indicate superior predictive performance.

\subsection{Implementation Details and Setups}
\label{setting}
We use a pre-trained CrossViT~\cite{chen2021crossvit} as our scale encoder. The small and large-scale branches have depths of 1 and 4, respectively, using patch sizes of 12 and 16, and token dimensions of 192 and 384. The number of attention heads is 6. For our frozen LLM module, we employ Llama-7B. We also include a standard transformer decoder with a depth of 1 for baseline comparisons. The model was trained for 9 epochs with a learning rate 2e-4 and a decay factor of 10 applied every 3 epochs. Batch sizes range from 16 to 128, depending on the dataset size. The dataset was divided into 80\% for training and 20\% for testing, and this process was repeated ten times to minimize performance bias. For other hyperparameters, we set $\lambda$ to 0.01, [$\alpha$, $\beta$] to [1/3, 3/4], $\gamma_1$ to 0.2, $\gamma_2$ to 0.7, and $\tau$ to 0.3. We report the median SRCC and PLCC to evaluate the model's prediction accuracy and monotonicity performance. 
All experiments were conducted on eight NVIDIA RTX 3090 GPUs.
\vspace{-0.3cm}

\subsection{Performance Comparison with SOTA}
As shown in Tab.~\ref{performance}, we compare our CSFIQA against 16 classical and state-of-the-art BIQA methods. For competing models, we used publicly available implementations or retrained them with official code. The comparison includes both hand-crafted feature methods (e.g., BRISQUE, ILNIQE) and deep learning approaches (e.g., LoDa, QFM-IQM).
CSFIQA outperforms all competing methods on six of the seven datasets, demonstrating robust performance across diverse image content and distortion types. The improvement is particularly stark on the challenging LIVEFB dataset, where our model achieves an 8.8\% performance gain. We attribute this success to our model's superior ability to distinguish quality information across different scales, a challenge prominent in the LIVEFB dataset. These results confirm the effectiveness and superiority of our method.

\subsection{Generalization Capability Validation}
To validate the generalization of our model, CSFIQA, we conducted cross-dataset experiments, training on one dataset and testing on others without any fine-tuning. As shown in Tab.~\ref{crossdata}, which reports the average SRCC scores, CSFIQA achieved the best performance in all evaluations, with particularly strong results on the LIVEC and KonIQ datasets.
We attribute this robust generalization to our model's unique architecture. The SCL and NSM modules excel at differentiating quality variations across scales, while the SFA module effectively focuses on the most salient information. This confirms our model's exceptional generalization ability.
\vspace{-5pt}
\subsection{Ablation study}
\noindent{\textbf{Overall.}} Tab.~\ref{ab_model} and Tab.~\ref{ab_guti} present the ablation performance of our proposed main framework, Scale Contrastive Learning (SCL) and Selective Focus Attention (SFA), along with their sub-modules on the LIVEC dataset. Our SCL framework primarily consists of Intra/Inter Scale Learning modules and a Noise Sample Matching (NSM) mechanism. Selective Focus Attention primarily consists of the Adaptive Filtering Selector (AFS) module for redundant information filtering and the Information Concentrator Module (ICM) for amplifying quality-relevant information. In Tab~\ref{ab_model}, we employ CrossViT (supplemented with a transformer decoder) as our baseline model. Tab~\ref{ab_guti} further refines the contribution of each sub-module to performance improvement.  Tab.~\ref{lda} and Tab.~\ref{range} present the performance of our hyperparameters on LIVEC and KonIQ datasets. The performance on the LIVE dataset is presented in Tab.~\ref{gamma} and Tab~\ref{tab:tau}. Notably, except for the observed hyperparameters, all other hyperparameters were selected according to the settings reported in Sec.~\ref{setting}.

\noindent{\textbf{Effect of Scale Contrastive learning Module (SCL).}}
Tab.~\ref{ab_model} demonstrates that SCL provides the most significant improvement to our method. This further illustrates SCL's effectiveness in mitigating the ``visual illusion'' problem. Specifically, inter/intra-SCL helps the model establish quality relationships between different images from varied perspectives, strengthening its quality perception capabilities across arbitrary scales. The NSM module is explicitly designed to address the ``visual illusion'' problem by bringing together features from different regions to distinguish subtle quality variations within the same image at different scales. The substantial improvement shown by the NSM module in Tab.~\ref{ab_guti} confirms this effectiveness.

\begin{table}[t]
\caption{Ablation experiments with different components on two datasets. The modules with the greatest improvement compared to the baseline are highlighted in gray.}
\centering
\small  
\setlength\tabcolsep{5pt}  
\renewcommand\arraystretch{1.15}  
\begin{tabular}{c|c|cc|cc} 
\toprule
\multirow{2}{*}{\textbf{Module}} & \multirow{2}{*}{\textbf{w/ Sub-Modules}} & \multicolumn{2}{c|}{\textbf{LIVEC}} & \multicolumn{2}{c}{\textbf{CSIQ}}  \\ 
\cmidrule{3-6}
                        &                          & \textbf{PLCC}  & \textbf{SRCC}            & \textbf{PLCC}  & \textbf{SRCC}             \\ 
\midrule
\multirow{3}{*}{w/ SCL}            & inter-SCL                    & 0.899 & 0.883           & 0.965 & 0.951            \\
& intra-SCL       & 0.902 & 0.884           & 0.965 & 0.954            \\
& \cellcolor{gray!25}\textbf{NSM}   & \cellcolor{gray!25}\textbf{0.903} & \cellcolor{gray!25}\textbf{0.887}          & \cellcolor{gray!25}\textbf{0.967} & \cellcolor{gray!25}\textbf{0.958}            \\
\midrule
\multirow{2}{*}{w/ SFA}     & \cellcolor{gray!25}\textbf{AFS}                 & \cellcolor{gray!25}\textbf{0.902} & \cellcolor{gray!25}\textbf{0.884}    & \cellcolor{gray!25}\textbf{0.967} & \cellcolor{gray!25}\textbf{0.953}           \\
& ICM   & 0.899 & 0.882    & 0.965 & 0.950          \\
\bottomrule
\end{tabular}
\label{ab_guti}
\end{table}

\begin{table}[t]
\caption{The ablation study about soft weight $\lambda$ in Eq.~\ref{loss}.}
\centering
\small  
\setlength\tabcolsep{6pt}  
\renewcommand\arraystretch{1.25}  
\begin{tabular}{c|cc|cc} 
\toprule
& \multicolumn{2}{c|}{\textbf{LIVEC}} & \multicolumn{2}{c}{\textbf{KonIQ}}   \\ 
\cmidrule{2-5}
\multirow{-2}{*}{\textbf{hyperparameter $\lambda$}} & \textbf{PLCC}  & \textbf{SRCC}           & \textbf{PLCC}  & \textbf{SRCC}             \\ 
\midrule
1 & 0.908 & 0.894 & 0.911 & 0.932 \\ 
0.1 & 0.915 & 0.900 & 0.917 & 0.940 \\
\rowcolor{gray!25}
\textbf{0.01} & \textbf{0.922} & \textbf{0.905} & \textbf{0.924} & \textbf{0.944} \\
0.001 & 0.910 & 0.887 & 0.922 & 0.941 \\
0.0001 & 0.909 & 0.893 & 0.918 & 0.938 \\
\bottomrule
\end{tabular}
\label{lda}
\end{table}

\begin{table}[t]
\caption{Ablation about the [$\alpha$, $\beta$] in AFS module.}
\centering
\small
\setlength\tabcolsep{5pt}
\renewcommand\arraystretch{1.35}  
\begin{tabular}{c|cc|cc} 
\toprule
& \multicolumn{2}{c|}{\textbf{LIVEC}} & \multicolumn{2}{c}{\textbf{KonIQ}}   \\ 
\cmidrule{2-5}
\multirow{-2}{*}{\textbf{range $[\alpha, \beta]$}} & \textbf{PLCC}  & \textbf{SRCC} & \textbf{PLCC}  & \textbf{SRCC} \\ 
\midrule
$[1/2]$ & 0.899 & 0.872 & 0.912 & 0.896 \\
$[1/6, 1/3]$ & 0.903 & 0.885 & 0.928 & 0.902 \\
$[1/5, 1/2]$ & 0.906 & 0.891 & 0.935 & 0.915 \\
$[1/4, 2/3]$ & 0.916 & 0.898 & 0.941 & 0.919 \\
\rowcolor{gray!25}
$[1/3, 3/4]$ & \textbf{0.922} & \textbf{0.905} & \textbf{0.944} & \textbf{0.924} \\
$[1/2, 4/5]$ & 0.918 & 0.899 & 0.938 & 0.916 \\
$[2/3, 1]$ & 0.908 & 0.887 & 0.927 & 0.902 \\
\bottomrule
\end{tabular}
\label{range}
\end{table}

\noindent{\textbf{Effect of Selective Focus Attention (SFA).}} Tab.~\ref{ab_model} demonstrates how the SFA module enhances quality assessment by removing redundant semantic information. This is further evidenced in Tab.~\ref{ab_guti}, where the AFS module shows greater performance gains compared to the ICM module. This further demonstrates that the performance gain from focusing becomes evident once redundant information is eliminated.

\noindent{\textbf{Effect of weight $\lambda$.}} We use $\lambda$ in Eq.~\ref{loss} to balance the scale contrastive learning. To this end, we conducted a sensitivity analysis on different values of $\lambda$ to investigate the effect of inter-scale contrastive learning. As shown in Tab.~\ref{lda}, we found that smaller values weaken inter-scale contrastive learning, while larger values cause excessive feature space changes and performance degradation, resulting in performance degradation. Therefore, we ultimately set $\lambda$=0.01.


\begin{figure*}[t]  
  \centering
  \includegraphics[width=0.8\textwidth]{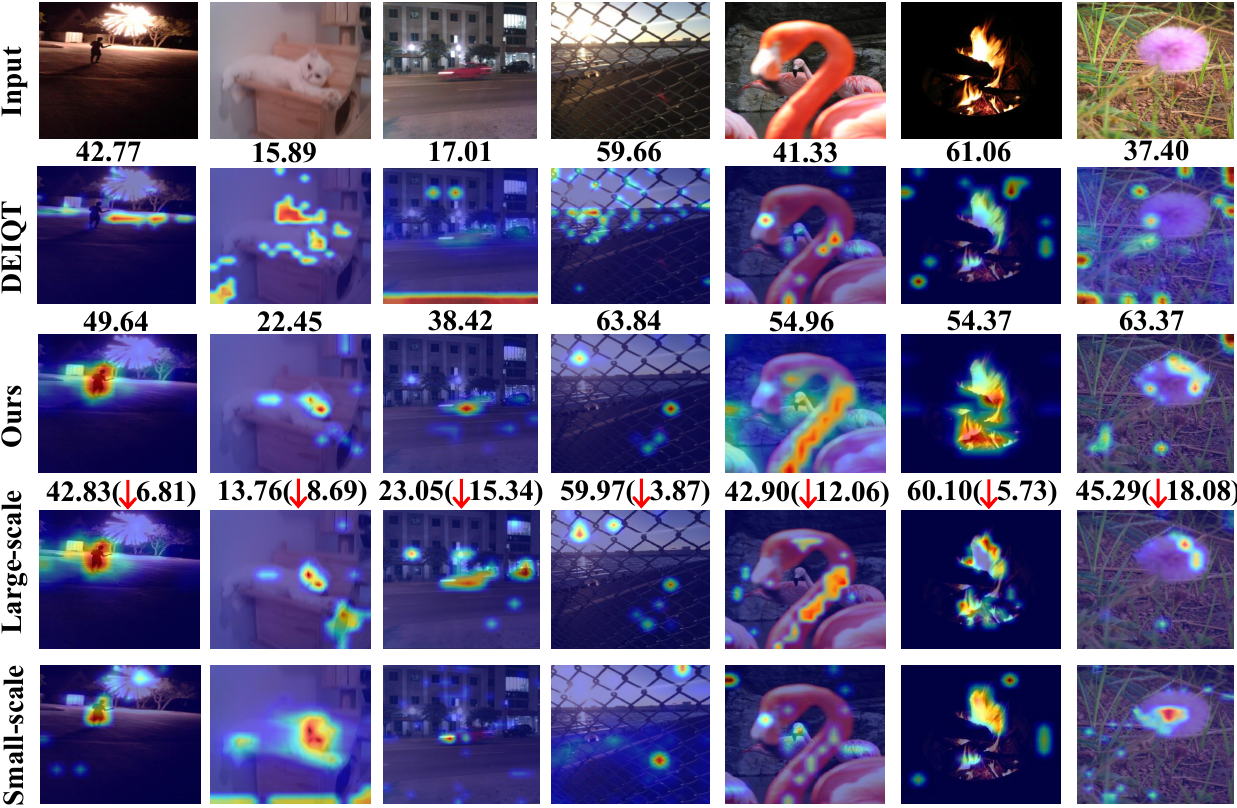}
  \caption{Grad-CAM Activation Maps of DEIQT and CSFIQA on LIVEC dataset. Scores below the first row indicate ground-truth MOS. Our model focuses more on distorted regions, leading to predictions closer to true values. Rows 1--3: input image, baseline CAM, CSFIQA CAM. Rows 4--5: large and small scale feature visualizations of CSFIQA.}
  \label{visual}
\end{figure*}

\begin{table}[t]
\caption{Impact of Ignoring Scale-Sensitive MOS Labels During Training: Comparison Between Different Scale / Same MOS and Different Scale / Different MOS Settings.}
\centering
\footnotesize  
\setlength\tabcolsep{5pt}  
\renewcommand\arraystretch{1.15}
\begin{tabular}{c|c|c|c|c} 
\toprule
Method & Different Scale & Different MOS & PLCC & SRCC \\
\midrule
CSFIQA & \CheckmarkBold & × & 0.710 & 0.641 \\
CSFIQA & \CheckmarkBold & \CheckmarkBold & 0.733 & 0.691 \\
\bottomrule
\end{tabular}
\label{mos_setting}
\end{table}

\begin{table}[t]
\centering
\tiny
\caption{Comparison of model parameters, MACs, and throughput.}
\resizebox{0.48\textwidth}{!}{
\begin{tabular}{lccc}
\midrule[0.5pt]  
\textbf{Method} & \textbf{Params} & \textbf{MACs} & \textbf{Throughput} \\ 
\midrule[0.5pt]
TReS~\citep{TReS} & 152.5M & 8.39G & 294(/s) \\ 
LoDa & 118.1M & 23.0G & 276(/s) \\ 
\rowcolor{gray!25}
Ours & 233.2M & 45G & 515(/s) \\ 
\midrule[0.5pt]  
\end{tabular}}
\label{cost}
\end{table}

\begin{table}[t]
\caption{The detailed params of CSFIQA.}
\centering
\small  
\setlength\tabcolsep{10pt}  
\renewcommand\arraystretch{1.0}  
\begin{tabular}{lccc}
\toprule
\textbf{Params} & \textbf{Total} & \textbf{Trainable} & \textbf{Frozen LLM} \\ 
\midrule
Ours & 233M & 31M & 202M \\ 
\bottomrule
\end{tabular}
\label{detailed_param}
\end{table}

\begin{table}[t]
\centering
\caption{Training time comparison of different methods.}
\small  
\setlength\tabcolsep{8pt}  
\renewcommand\arraystretch{1.3}  
\begin{tabular}{lcc}
\toprule
\textbf{Method} & \textbf{Per Epoch} & \textbf{Total Time} \\ 
\midrule
HyperIQA~\citep{hypernet} & 928s & 11861s \\ 
LoDa~\citep{xu2024boosting} & 586s & 5798s \\ 
\rowcolor{gray!25}
\textbf{Ours} & \textbf{270s} & \textbf{2724s} \\ 
\bottomrule
\end{tabular}
\label{tab:training_time}
\vspace{-0.2cm}
\end{table}

\begin{table}[t]
\caption{Ablation Study on Training Time for the AFS Module.}
\centering
\small  
\setlength\tabcolsep{12pt}  
\renewcommand\arraystretch{1.0}  
\begin{tabular}{lcc}
\toprule
\textbf{Time} & \textbf{w/ AFS} & \textbf{w/o AFS}  \\ 
\midrule
Ours & 2724s & 8031s  \\ 
\bottomrule
\end{tabular}
\label{tab:time}
\vspace{-0.5cm}
\end{table}

\begin{table}[t]

\centering
\caption{The performance of CSFIQA in terms of SRCC on the LIVE dataset with different values of [$\gamma_1$,$\gamma_2$].}
\normalsize
\setlength\tabcolsep{4pt}
\renewcommand\arraystretch{1.4}
\begin{tabular}{lccccc}
\toprule
\textbf{[$\gamma_1$,$\gamma_2$]} & [0.1,0.5] & [0.1,0.8] & \textbf{[0.2,0.7]} & [0.2,0.8] & [0.3,0.9] \\ 
\midrule
CSFIQA & 0.897 & 0.903 & \textbf{0.905} & 0.900 & 0.883 \\ 
\bottomrule
\end{tabular}
\label{gamma}
\end{table}

\begin{table}[t]
\caption{The performance of CSFIQA in terms of SRCC on the LIVE dataset with different values of $\tau$. The best performance is achieved when $\tau$ is set to 0.3. Therefore, we set the hyperparameter $\tau$ to 0.3.}
\centering
\small
\setlength\tabcolsep{5pt}
\renewcommand\arraystretch{1.5}
\begin{tabular}{cccccc}
\toprule
$\boldsymbol{\tau}$ & 0.1 & 0.2 & \textbf{0.3} & 0.4 & 0.5  \\ 
\midrule
CSFIQA & 0.902 & 0.901 & \textbf{0.905} & 0.904 & 0.900  \\ 
\bottomrule
\end{tabular}
\label{tab:tau}
\vspace{-0.4cm}
\end{table}

\begin{figure}[htbp]
  \centering
\includegraphics[width=0.47\textwidth]{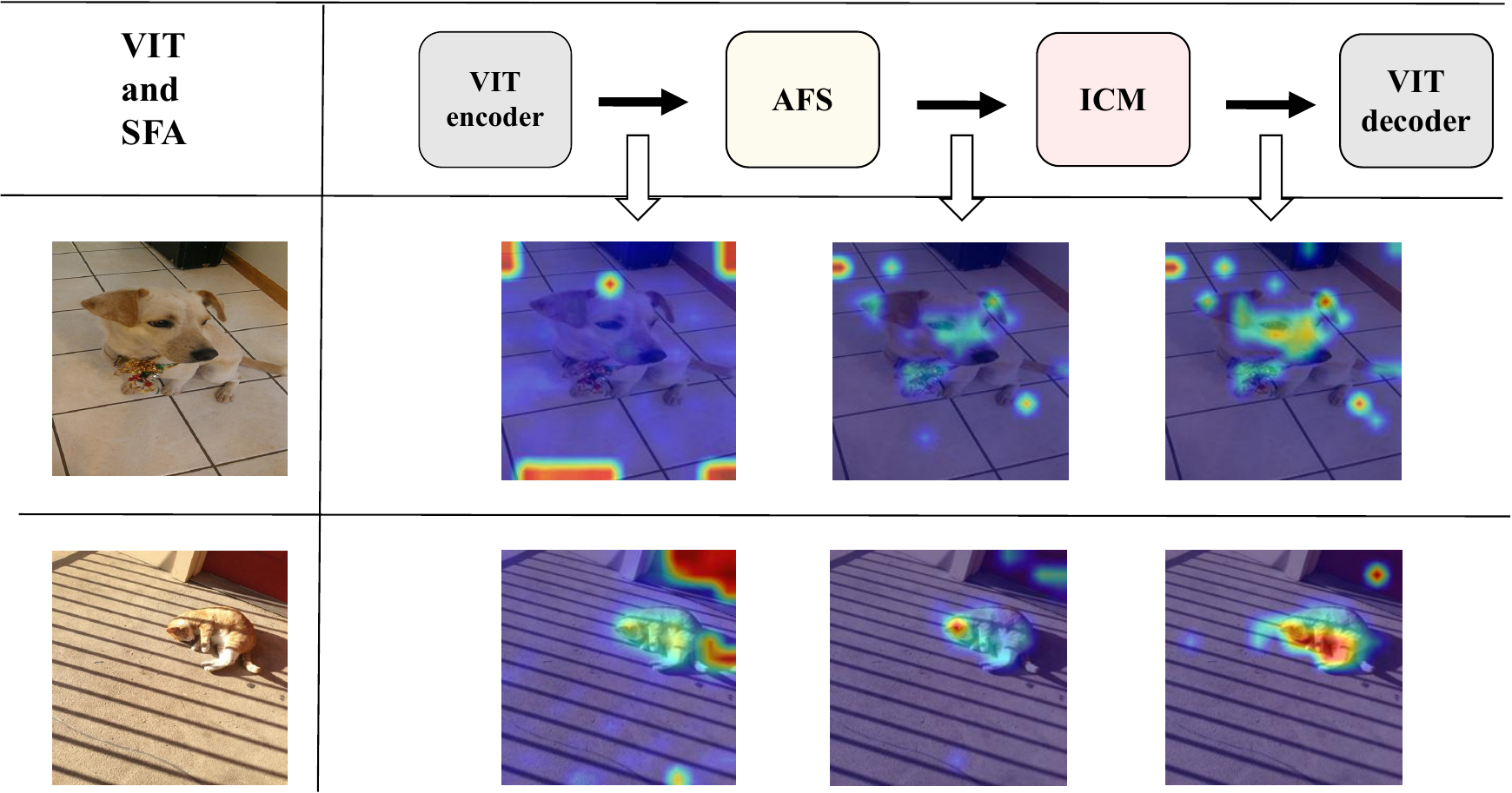}
  \caption{ 
We conducted a visualization ablation experiment on the SFA module, where we obtained the attention map of each layer before entering the module using single Grad-CAM, in order to analyze the role played by the selection and focusing modules.
  }
  \label{attention_fulu}
  \vspace{-0.4cm}
\end{figure}

\noindent{\textbf{Effect of weight $[\alpha, \beta]$.}} Tab.~\ref{range} presents our ablation study on the AFS module's filtering range $[\alpha, \beta]$. 
The choice of $k$ is critical for performance; we found that using a single, fixed value for $k$ resulted in instability. To enhance robustness, we instead sample $k$ from a learnable range $[\alpha, \beta]$. The results reveal a clear trade-off. A range set too low results in insufficient information aggregation, causing a sharp performance decline. Conversely, a range set too high introduces semantically irrelevant information, which also degrades performance. We achieved optimal results with $[\alpha, \beta]$ set to \textbf{[1/3, 3/4]}, confirming this range provides the best balance of focused yet sufficient information.
\subsection{Qualitative Analysis} 
\noindent{\textbf{Feature Visualization.}} 
Fig.~\ref{visual} presents GradCAM~\cite{grad} visualizations comparing the feature attention of our model, CSFIQA, against the baseline. 
The results clearly show that CSFIQA accurately focuses on distorted image regions and accurately predict the quality score.
In contrast, the baseline is often distracted by irrelevant content, a phenomenon we term ``visual illusion'', which impairs its judgment.
This improved focus stems from our model's ability to effectively process and integrate quality information from different scales, an ability deliberately cultivated during training.
Notably, the last two rows display our model's visualization at each distinct scale. 
In these ``visual illusion'' cases, the perceived quality regions for the same image vary significantly across scales, leading to a large discrepancy in their quality assessments. 
This phenomenon aligns perfectly with our research motivation.

\noindent{\textbf{Scale Quantitative Analysis.}} To investigate the impact of scale variations on image quality assessment, we conducted experiments using two settings in the CSFIQA framework, as shown in Tab.~\ref{mos_setting}. The first setting, Different Scale / Same MOS, assigns the same quality label across varying scales, while the second setting, Different Scale / Different MOS, assigns distinct quality labels based on scale differences. The results, evaluated on the LiveFB dataset, show that the Different Scale / Different MOS setting outperforms Different Scale / Same MOS, emphasizing the importance of incorporating scale-aware features in quality prediction.


\subsection{Visualizations of the SFA module} 
\label{sec:SFA_detail}
The SFA module mainly consists of the Adaptive Filtering Selector (AFS) and the Information Concentrator Module (ICM). Our ICM structure is straightforward, consisting of only two linear layers and a frozen Llama-7B block. The first linear layer maps the visual features to the same dimension as Llama-7B, while the second linear layer maps it back to the original feature dimension. Both of these linear layers are trainable. After Filtering Attention, important visual features are aligned under the Llama module, rich in prior knowledge, achieving a focused effect. We further validate our approach by visualizing each layer in Fig.~\ref{attention_fulu}.

It can be observed that before entering the AFS module, the model focuses on non-target areas, resulting in a visual illusion. After passing through the AFS module, the model selects and filters the main target features, removing most of the redundant information. Then, the ICM module focuses on the key target features. Due to the absence of the SCL and NSM modules, some noise still remains in the final result.

\subsection{The computational cost details of CSFIQA}
The following four tables respectively show the training time (Tab.~\ref{tab:training_time}) and inference speed (Tab.~\ref{cost}) of CSFIQA compared with multiple SOTA methods on the Koniq dataset (Batch=64), the parameter composition of CSFIQA modules (Tab.~\ref{detailed_param}), and the ablation experiments conducted on the filtering mechanism in the SFA module (Tab.~\ref{tab:time}).

Our initial intention was not to improve model speed, but to reduce the impact of information redundancy on model quality assessment. Based on your valuable suggestion, we tested our computational cost details of CSFIQA. Although our parameter count is nearly double that of Loda, we have significantly reduced computational costs thanks to AFS (filtering mechanism) 's handling of information redundancy. When we removed the AFS module, the training time increased by nearly 3 times (Tab.~\ref{tab:time}). Notably, our large parameter count is mainly due to the frozen LLM module, with only 31M learnable parameters (Tab.~\ref{detailed_param}).

\subsection{Further Ablation Studies on Hyperparameters}
We further report ablation studies on the hyperparameters $\gamma_1$ and $\gamma_2$ in Eq.1, as well as the temperature hyperparameter $\tau$ in Eq.3. Our hyperparameters $\gamma_1$ and $\gamma_2$ are used for acquiring positive and negative samples, respectively. In Tab.~\ref{gamma}, we report the performance of our model on LIVEC with different hyperparameter settings. The results indicate that our model performs best when $\gamma_1$ is set to 0.2 and $\gamma_2$ to 0.7. Similarly, we report the SRCC performance on LIVEC with different temperature $\tau$  in Tab.~\ref{tab:tau}, with our model achieving optimal results when the temperature is set to 0.3.

\section{Conclusion}\label{section4}
This paper introduces LML-IQA, an innovative No-Reference Image Quality Assessment (NR-IQA) model grounded in local manifold learning. Central to this approach is the strategic use of visual saliency cropping and intra-class negative classes. 
This ensures a consistent representation, while simultaneously avoiding excessive proximity to prevent local manifold collapse.
In parallel, inter-class negative classes are utilized to partition the overall manifold structure, ensuring a robust and comprehensive image quality assessment.
Furthermore, we establish a mutual learning paradigm between the teacher and student through the EMA algorithm.
This approach facilitates adaptive adjustments in visual saliency cropping, maintaining the stability of local manifolds during the learning process.
Experimental results conducted on eight IQA datasets highlight the superiority of our proposed approach.




\bibliographystyle{unsrt}
\bibliography{ref}

\vfill

\end{document}